\documentclass{article}

\usepackage{arxiv}

\usepackage{algorithm}

\usepackage{algpseudocode}
\usepackage{amsmath,amssymb,epsfig,multirow,amsthm,array,bm}

\usepackage[utf8]{inputenc} 
\usepackage[T1]{fontenc}    
\usepackage{hyperref}       
\usepackage{url}            
\usepackage{booktabs}       
\usepackage{amsfonts}       
\usepackage{nicefrac}       
\usepackage{microtype}      
\usepackage{lipsum}		
\usepackage{graphicx}
\usepackage{natbib}
\usepackage{doi}

\usepackage{caption}
\usepackage{subcaption}

\usepackage{framed}
\usepackage{ulem}
\usepackage{siunitx}
\usepackage{fancyvrb}

\newcommand{\Beta}{{\mathcal{B}}}

\newcommand{\Bernoulli}{{\mathcal{B}ern}}

\renewcommand{\a}{{a}}
\newcommand{\deltab}{{\bm{\delta}}}
\newcommand{\thetab}{{\bm{\theta}}}

\newcommand{\EE}{{\mathbb{E}}}

\AtBeginDocument{\let\latexlabel\label}

\DefineVerbatimEnvironment{Sinput}{Verbatim}{fontshape=sl}
\DefineVerbatimEnvironment{Soutput}{Verbatim}{}
\DefineVerbatimEnvironment{Scode}{Verbatim}{fontshape=sl}

\DefineVerbatimEnvironment{Code}{Verbatim}{}
\DefineVerbatimEnvironment{CodeInput}{Verbatim}{fontshape=sl}
\DefineVerbatimEnvironment{CodeOutput}{Verbatim}{}
\newenvironment{CodeChunk}{}{}

\title{$\beta^{4}$-IRT: A New  $\beta^{3}$-IRT with Enhanced Discrimination Estimation}

\author{ Manuel {Ferreira-Junior} \\
	Departmento de Estat\'istica\\
	Universidade Federal da Para\'iba\\
	Jo\~ao Pessoa, Brazil\\
	\texttt{ferreira.jr.ufpb@gmail.com} \\
	\And
	J\'essica T.S. Reinaldo \\
	Centro de Inform\'atica\\
	Universidade Federal de Pernambuco\\
	Recife, Brazil \\
	\texttt{jtsr@cin.ufpe.br} \\
        \And
	Telmo M. {Silva Filho} \\
	Department of Engineering Maths\\
	University of Bristol\\
	Bristol, United Kingdom \\
	\texttt{telmo.silvafilho@bristol.ac.uk} \\
        \And
        Eufr\'asio A. {Lima Neto} \\
	Departmento de Estat\'istica\\
	Universidade Federal da Para\'iba\\
	Jo\~ao Pessoa, Brazil\\
	\texttt{eufrasio@de.ufpb.br} \\
	\And
	Ricardo B.C. Prud\^encio \\
	Centro de Inform\'atica\\
	Universidade Federal de Pernambuco\\
	Recife, Brazil \\
	\texttt{rbcp@cin.ufpe.br} \\
}

\renewcommand{\shorttitle}{$\beta^{4}$-IRT: A New  $\beta^{3}$-IRT with Enhanced Discrimination Estimation}

\hypersetup{
pdftitle={B4IRT: A New BIRT with Better Discrimination Estimation},
pdfsubject={q-bio.NC, q-bio.QM},
pdfauthor={Manuel Ferreira Junior, Jessica T.S.~Reinado, Telmo M.~Silva Filho, Eufrasio A.~Lima Neto, Ricardo B.C.~Prudencio},
pdfkeywords={Item response theory, Latent variable models, Discrimination estimation, Python package},
}

\begin{document}
\maketitle

\begin{abstract}
	Item response theory aims to estimate respondent's latent skills from their responses in tests composed of items with different levels of difficulty. Several models of item response theory have been proposed for different types of tasks, such as binary or probabilistic responses, response time, multiple responses, among others. In this paper, we propose a new version of $\beta^3$-IRT, called $\beta^{4}$-IRT, which uses the gradient descent method to estimate the model parameters. In $\beta^3$-IRT, abilities and difficulties are bounded, thus we employ link functions in order to turn $\beta^{4}$-IRT into an unconstrained gradient descent process. The original $\beta^3$-IRT had a symmetry problem, meaning that, if an item was initialised with a discrimination value with the wrong sign, e.g. negative when the actual discrimination should be positive, the fitting process could be unable to recover the correct discrimination and difficulty values for the item. In order to tackle this limitation, we modelled the discrimination parameter as the product of two new parameters, one corresponding to the sign and the second associated to the magnitude. We also proposed sensible priors for all parameters. We performed experiments to compare $\beta^{4}$-IRT and $\beta^3$-IRT  regarding parameter recovery and our new version outperformed the original $\beta^3$-IRT. Finally, we made $\beta^{4}$-IRT publicly available as a Python package, along with the implementation of $\beta^3$-IRT used in our experiments.
\end{abstract}

\keywords{Item response theory \and Latent variable models \and Discrimination estimation \and Python package}

\section[Introduction]{Introduction} \label{sec:intro}

Item Response Theory (IRT) is widely adopted in the field of psychometrics to estimate latent abilities of human test respondents. Unlike classical test theory, which assesses performance at the test level, IRT focuses on items and aims to model responses given by respondents of different abilities to items of different difficulties, both measured on a known scale \citep{embretson2013item}. The concept of an item depends on the application and can represent, for example, exam, open-ended or multiple choice questions. In practice, IRT models estimate latent skills and difficulties based on responses observed in a test and have been commonly applied to measure student performance on exams.

There are different IRT models in literature, with respect to the range of responses. In this paper we focus on the 
$\beta^{3}-$IRT model \citep{chen2019birt}, which considers bounded continuous responses, suitable to model, for instance, success rates and probabilities.
The $\beta^{3}-$IRT model is more flexible than other continuous IRT models since it can result on Item Characteristic Curves (ICCs) that are not
limited to logistic curves. ICCs with different shapes (e.g., sigmoid, parabolic and anti-sigmoid) can be obtained, which is more flexible to fit responses for different items.

The original $\beta^3$-IRT model, 
as proposed by \cite{chen2019birt}, has a symmetry problem meaning that, for a respondent with a certain ability value, two items, one with low difficulty and positive discrimination and another with high difficulty and negative discrimination, could have the same expected response. As a result, if an item was initialised with with the wrong sign for its discrimination value, the fitting process could be unable to recover the correct discrimination value. This issue is not exclusive to $\beta^3$-IRT and is associated to any IRT model which considers a discrimination parameter. Additionally, the code that is available online\footnote{https://github.com/yc14600/beta3\_IRT}, which performs a variational inference-based process to estimate the full posterior distributions of its parameters uses a Python 2 library that is now obsolete.

Thus, in this paper we improve $\beta^3$-IRT in a few ways. First, we tackle the symmetry limitation by modelling the discrimination parameter using the multiplication of two new values, one corresponding to the sign and the second associated to the magnitude. These new parameters are kept fixed for the first fitting iterations, in order to better estimate abilities and difficulties. After these first iterations, the discrimination parameters are optimised along with abilities and difficulties. We also provide sensible priors for abilities, difficulties and both discrimination parameters. Together, this two-step optimisation, the factoring of discrimination into two parameters and the suggested priors help us to avoid the symmetry problem, improving the parameter estimates. Additionally, our improved $\beta^3$-IRT, which is called $\beta^{4}$-IRT, uses gradient descent to estimate the model parameters. This allows us to leverage cutting-edge Python libraries for fast GPU-based computation. In $\beta^3$-IRT, abilities and difficulties are bounded in $(0, 1)$, thus we employ link functions in order to formulate $\beta^{4}$-IRT as an unconstrained gradient descent process.

We perform an experimental analysis of $\beta^{4}$-IRT and $\beta^3$-IRT regarding parameter recovery, i.e. how well a fitted model estimates the original parameter values used to produce an artificial response dataset. Finally, this work provides a publicly available Python library for $\beta^{4}$-IRT.

The paper is organised as follows: Section \ref{sec:irt} discusses  the $\beta^3$-IRT model in detail and explains its limitations; Section \ref{sec:birtgd} presents the mathematical definition for the new $\beta^{4}$-IRT model as well as the algorithm for parameter estimation; Section \ref{sec:experiments} provides an experimental analysis of parameter recovery and computing time; and finally, Section \ref{sec:summary} brings some final remarks.

\section[Item response theory]{Item response theory}
\label{sec:irt}






An IRT model assumes that, for each item $j$, a respondent $i$ produces a response that is a function of the respondent's ability and the item's difficulty, sometimes including other parameters for the items, such as discrimination and guessing.

Most works on IRT assume that the response $x_{ij}$ is binary, which is usually encoded as $x_{ij}=1$ if the $j$-th item was correctly answered by the $i$-th respondent, otherwise $x_{ij}=0$ \citep{bachrach2012grade,embretson2013item,martinez2016making,twomey2022equitable}. These models commonly assume that a response $x_{ij}$ follows a Bernoulli distribution with probability of success $p_{ij}$ defined as a logistic function of the respondent's latent ability $\theta_i$ and of two latent parameters associated to each item, the difficulty $\delta_j$ and the discrimination $a_j$, as given by Equation (\ref{eq:genIRT}):

\begin{equation} 
\label{eq:genIRT}
    x_{ij} = \Bernoulli(p_{ij}),\; 
    p_{ij} = \sigma(-{a_j} d_{ij}),\;
    d_{ij} = {\theta_i-\delta_j},\\ 
\end{equation}

\noindent where $\sigma(\cdot)$ is the logistic function, with location parameter $\delta_j$ and shape parameter $a_j$,  $j=1,\ldots,N$ and $i=1,\ldots,M$. This model, known as 2-parameter logistic IRT (2PL-IRT) results in an item characteristic curve (ICC) that maps ability to expected response as shown in  Equation (\ref{eq:2plIRT}):

\begin{equation}
\label{eq:2plIRT}
\EE[x_{ij}|\theta_i,\delta_j,a_j] = p_{ij} = \frac{1}{1+e^{-a_j(\theta_i-\delta_j)}}.
\end{equation}

\noindent When $\theta_i = \delta_j$, the expected response is $0.5$. Moreover, if $a_j=1, \forall j=1,\ldots,N$, a simpler model is obtained, known as 1PL-IRT, which describes the items only by their difficulties. In general, the discrimination $a_j$ indicates how the probability of correct answers changes as skill increases. High discriminations induce steep ICCs at the point where skill equals difficulty, with small changes in skill causing large changes in the probability of correct answer.


Despite their extensive use in psychometry, binary IRT models have limited use when res\-ponses are produced on continuous scales. In particular, binary models are not suitable if the evaluated responses are estimates of probabilities or proportions, as in the case explored by \cite{chen2019birt}, where each student could respond to the same item multiple times and IRT was used to model the proportion of times that the student was correct for each item. For such cases, a different model, called $\beta^3$-IRT was proposed by \cite{chen2019birt}. Equation (\ref{eq:model_def}) defines $\beta^3$-IRT, where $p_{ij}$ is the observed response of respondent $i$ for item $j$, which is assumed to follow a Beta distribution with parameters $\alpha_{ij}$ and $\beta_{ij}$ defined as functions of the respondent's ability $\theta_i$ and of the item's difficulty $\delta_j$ and discrimination $a_j$:

\begin{figure*}[!ht]
\centering
\begin{subfigure}{0.45\linewidth}
\includegraphics[width=\linewidth,trim={1cm 0.cm 0.cm 0.cm}]{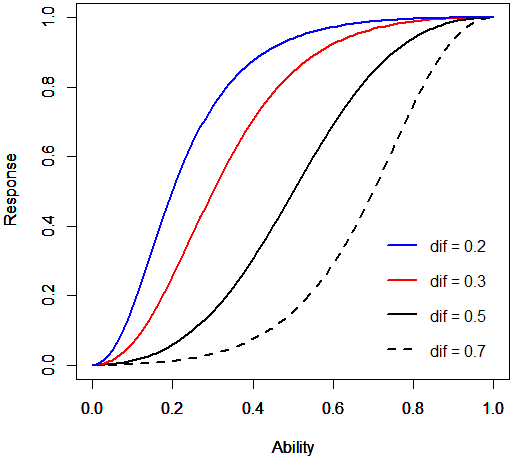}
\caption{$\a_j = 2$.}
\label{fig:ICC_discr2}
\end{subfigure}
\quad
\begin{subfigure}{0.45\linewidth}
\includegraphics[width=\linewidth,trim={1cm 0.cm 0.cm 0.cm}]{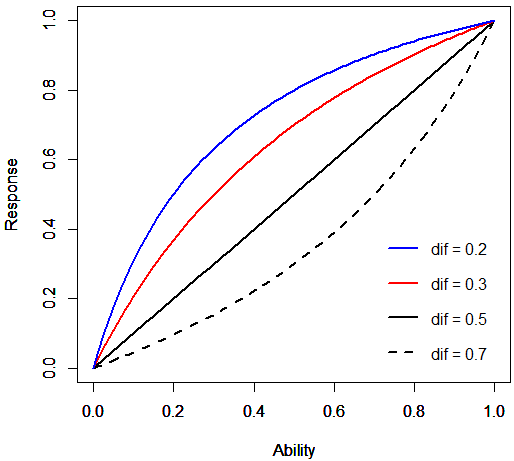}
\caption{$\a_j = 1$.}
\label{fig:ICC_discr1}
\end{subfigure}
\\
\begin{subfigure}{0.45\linewidth}
\includegraphics[width=\linewidth,trim={1cm 0.cm 0.cm 0.cm}]{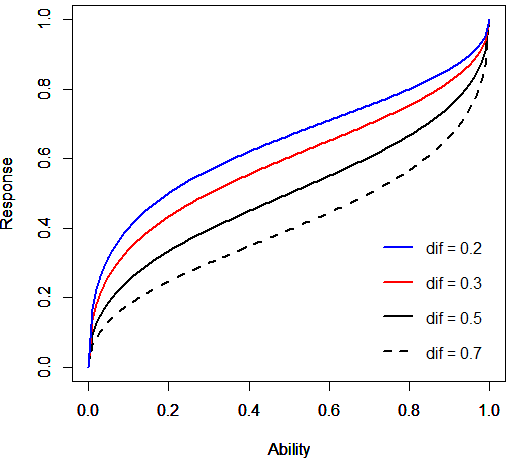}
\caption{$\a_j = 0.5$.}
\label{fig:ICC_discr05}
\end{subfigure}
%
\caption{Examples of $\beta^3$-IRT ICCs for different values of difficulty and discrimination. Steeper ICCs result of higher discrimination values, while difficulty determines the ability needed to surpass a response of $0.5$. Source: \cite{chen2019birt}.}
\label{fig:betaICCs}
\end{figure*}

\begin{align} \label{eq:model_def}
    p_{ij} & \sim \Beta(\alpha_{ij}, \beta_{ij}), \notag \\
    \alpha_{ij} &= 
    \left(\frac{\theta_i}{\delta_j}\right)^{a_j}, \notag 
    \beta_{ij} = 
    \left(\frac{1-\theta_i}{1-\delta_j}\right)^{a_j}, \notag \\
    \theta_i &\sim \Beta(1,1),\;
    \delta_j \sim \Beta(1,1),\;
    a_j \sim \mathcal{N}(1, \sigma^2_{0}).
\end{align}

Here, $\sigma^2_{0}$ is a hyperparameter of the model, which the authors set as $1$ in their experiments. In this model, the ICC is defined by the expected value of $\Beta(\alpha_{ij}, \beta_{ij})$, taking the form given by Equation (\ref{eq:ep_birt}):

\begin{equation}\label{eq:ep_birt}
    \begin{split}
        \EE[p_{ij}|\theta_i,\delta_j,a_j] &= \frac{\alpha_{ij}}{\alpha_{ij}+\beta_{ij}} 
        = \frac{1}{1+\left(\frac{\delta_j}{1-\delta_j}\right)^{a_j} \left(\frac{\theta_i}{1-\theta_i}\right)^{-a_j}}.
    \end{split}
\end{equation}

This parametrisation enables $\beta^3$-IRT to obtain non-logistic ICCs, with the difficulty $\delta_j$ as a location parameter, similarly to logistic IRT models. The response is 0.5 when $\theta_i=\delta_j$ and the curve has slope $\a_j/(4\delta_j(1-\delta_j))$ at that point. Figure \ref{fig:betaICCs} shows examples of $\beta^3$-IRT ICCs with different shapes, depending on $a_j$. For $a_j > 1$, we see a sigmoid shape, similar to logistic IRT models; $a_j = 1$ gives parabolic curves, with vertex at $0.5$; and $0 < a_j < 1$ leads to an anti-sigmoidal behaviour. The model also allows for negative discriminations. In such cases, $-1 < a_j < 0$ and $a_j < -1$ give decreasing anti-sigmoid and decreasing sigmoid ICCs, respectively.


Note that correctly estimating the discrimination parameter, particularly its sign, is crucial, as it encodes information about the perceived behaviour of an item. A sigmoidal ICC means that the item is good at discriminating respondents in the middle of the ability range, while an anti-sigmoidal one does a good job at detecting different abilities in the low and high ranges. Additionally, negative discriminations could be interpreted as corresponding to items that are harder for respondents with higher abilities. Thus, \cite{chen2019birt} use negative discriminations to identify `noisy' items.

\cite{chen2019birt} tested two inference methods for $\beta^3$-IRT, one was conventional Maximum Likelihood (MLE), using the likelihood function shown in Equation (\ref{eq:model_def}). The second method was Bayesian Variational Inference (VI) \citep{bishop2006pattern}, which they applied to their experiments with IRT to evaluate machine learning classifiers. 

Independently of the inference method, $\beta^3$-IRT is a highly non-identifiable model, because of its symmetry \citep{nishihara2013detecting}, which can result in undesirable combinations of the latent variables. For instance, when $p_{ij}$ is close to $1$, it usually indicates $\alpha_{ij} > 1$ and $\beta_{ij} < 1$, which can arise either from $\thetab_i > \deltab_j$ with positive $\a_j$, or from $\thetab_i < \deltab_j$ with negative $\a_j$. 

To show the impact of this non-identifiability on parameter estimation, we sampled 1000 abilities, difficulties and discriminations from the priors defined in Equation (\ref{eq:model_def}), setting $\sigma_0^2 = 1$. Then, for each $i$-th respondent and $j$-th item, we generated the response $p_{ij}$ by taking the mean of $100$ samples from the corresponding $\Beta(\alpha_{ij}, \beta_{ij})$ distribution. Then, we fit a $\beta^3$-IRT model using MLE (this implementation is available as part of the Python package we present in Section \ref{sec:birtgd}). Here, we train the $\beta^3$-IRT model using 50,000 iterations.

\begin{figure*}[!htb]
\centering
\includegraphics[width=\linewidth,trim={0cm 0.cm 12.cm 0.cm},clip]{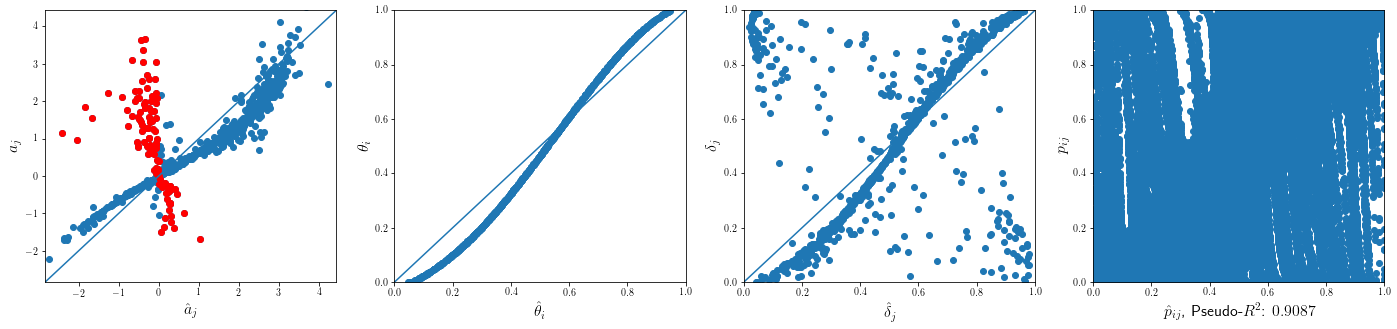}
%
\caption{Scatter plots showing sampled discriminations (left), abilities (centre) and difficulties (right) used to generate a $1000\times1000$ response matrix, and their estimates produced by $\beta^3$-IRT. Red dots on the discrimination plot represent discriminations estimated with flipped signs (109 out of 1000 discriminations). }
\label{fig:b3-0-init-1000x1000(109)}
\end{figure*}

Figure \ref{fig:b3-0-init-1000x1000(109)} shows the original parameter values and their estimates. The discriminations and their estimates seem to follow an inverted-sigmoidal relationship, moving away from the diagonal for lower and higher values. Additionally, the 109 red dots represent discriminations that were estimated with the wrong sign. These were likely initialised with flipped signs and, due to the symmetry of the model, ended up pushing their corresponding difficulties away from their target values, which shows as an orbit around the diagonal in the difficulty plot. Finally, due to not fitting certain discriminations and difficulties correctly, the estimated abilities were also pushed away from their original values.

Some attempts can be made to avoid this problem. In their VI implementation, \cite{chen2019birt} updated discrimination as a global variable after ability and difficulty converged at each step. They also set the prior of discrimination as $\mathcal{N}(1, 1)$ to reflect the assumption that discrimination is more often positive than negative. In our implementation, we can set a number of initial iterations, say 1000, where we keep all discriminations fixed at $\hat{a}_j=1$ and optimise only abilities and discriminations. Then we allow the discriminations to be optimised as well. This has a positive impact, reducing the number of flipped discrimination signs to 37, but does not definitely solve the problem. In the next Section we present a new model based on $\beta^3$-IRT, which introduces a new parameter to estimate the signs of the discriminations, leading to better parameter estimates.

\begin{figure*}[!htb]
\centering
\includegraphics[width=\linewidth,trim={0cm 0.cm 12.cm 0.cm},clip]{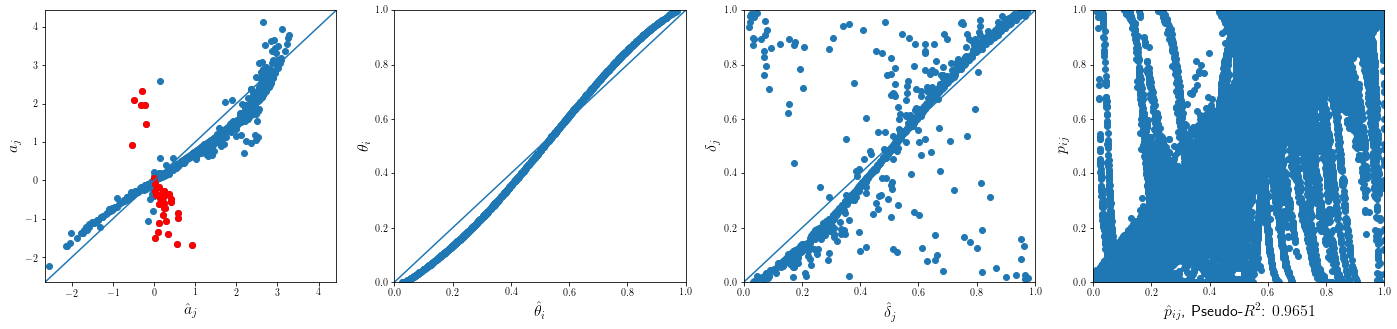}
%
\caption{Scatter plots showing sampled discriminations (left), abilities (centre) and difficulties (right) used to generate a $1000\times1000$ response matrix, and their estimates produced by $\beta^3$-IRT with 1000 initial iterations with fixed discriminations. Red dots on the discrimination plot represent discriminations estimated with flipped signs (37 out of 1000 discriminations). }
\label{fig:b3-1000-init-1000x1000(37)}
\end{figure*}

\section[B4IRT: Mathematical definition and implementation]{$\beta^{4}-$IRT: Mathematical definition and implementation}\label{sec:birtgd}


As mentioned in the previous section, $\beta^{3}-$IRT is sometimes unable to overcome a poor initialisation of its discriminations. Motivated by this limitation, we propose the novel $\beta^{4}-$IRT model, whose ICC is given by Equation (\ref{eq:fo}):


\begin{equation}
    \label{eq:fo}
    E[p_{ij} | \theta_i, \delta_j, \omega_j, \tau_j] = \frac{1}{1 + \big(\frac{\delta_j}{1 - \delta_j}\big)^{\tau_j\cdot\omega_j }\cdot\big(\frac{\theta_i}{1 - \theta_i}\big)^{-\tau_j\cdot\omega_j }}.
\end{equation}


Equation (\ref{eq:fo}) substitutes the discrimination $a_j$ in Equation (\ref{eq:ep_birt}) with the product of two new parameters, $\omega_j$ and $\tau_j$, which represent the sign and the absolute value of the discrimination, respectively. This decomposition of the discrimination parameter aims to reduce the symmetry problem, as the direction and the magnitude of the discrimination will be  optimised separately. 



As seen in Equation (\ref{eq:model_def}) for the  $\beta^{3}-$IRT model, parameters $\theta_i$ and  $\delta_j$ take values from $(0, 1)$, while the discrimination $a_j$ has infinite support. In the new model, $\theta_i$ and $\delta_j$ have kept their supports, while $\tau_j$ and $\omega_j$ take their values from $(-1, 1)$ and $(0, \infty)$, respectively. In order to improve the optimisation process, the constraints on minimum and maximum values were removed by adopting link functions. Thus, the gradient descent method in $\beta^{4}-$IRT does not update the values of the four parameters directly. Instead, we introduce four new parameters ($t_i$, $d_j$, $b_j$ and $o_j$) with values in $\mathbb{R}$, which are used  to estimate the original parameters by way of link functions, as follows:



\begin{align*}
\refstepcounter{equation}\latexlabel{eq:theta:t}
\refstepcounter{equation}\latexlabel{eq:delta:d}
\refstepcounter{equation}\latexlabel{eq:omega:o}
\refstepcounter{equation}\latexlabel{eq:beta:b}
\theta_i = \sigma(t_i) = \frac{1}{1 + e^{-t_i}}, &\qquad \delta_j = \sigma(d_j),
\tag{\ref*{eq:theta:t}, \ref*{eq:delta:d}}\\
\omega_j = \text{softplus}(o_j) = \text{ln}(1 + e^{o_j}), &\qquad \tau_j = \text{tanh}(b_j) = \frac{e^{b_j} - e^{-b_j}}{e^{b_j} + e^{-b_j}}.
\tag{\ref*{eq:omega:o}, \ref*{eq:beta:b}}
\end{align*}


The estimation of the $t_i$,  $d_j$, $o_j$ and $b_j$ 
in 
$\beta^{4}-$IRT is carried out using the gradient descent method, by minimising the cost function given by Equation (\ref{eq:crossentropy}):


\begin{equation}
    \label{eq:crossentropy}
    H = -\sum_{i=1}^{M}\sum_{j=1}^{N} p_{ij}\cdot \ln(\hat{p}_{ij}),
\end{equation}

where $\hat{p}_{ij}$ is the estimated response and is calculated using Equations (\ref{eq:fo}), (\ref{eq:theta:t}), (\ref{eq:delta:d}), (\ref{eq:omega:o}) and (\ref{eq:beta:b}).
The partial derivatives of $H$ with regards to $d_j$, $t_i$,  $o_j$ and $b_j$ are given by Equations (\ref{eq:d1}), (\ref{eq:d2}), (\ref{eq:d3}) and (\ref{eq:d4}), respectively:
\begin{align}
    \label{eq:d1}
    &\frac{\partial H}{\partial d_j} =  \sum_{i=1}^{M}\sum_{j=1}^{N} p_{ij}\cdot w_j\cdot \tau_j \cdot \Delta(\delta_j) \cdot \Phi(\theta_i, \delta_j)^{w_j\cdot \tau_j} \cdot \hat{p}_{ij}\cdot e^{-d_j}\cdot \sigma(d_j)^{2},\\
    \label{eq:d2}
    &\frac{\partial H}{\partial t_i} = - \sum_{i=1}^{M}\sum_{j=1}^{N} p_{ij}\cdot w_j\cdot \tau_j \cdot \Theta(\theta_i) \cdot \Phi(\theta_i, \delta_j)^{w_j\cdot \tau_j} \cdot \hat{p}_{ij} \cdot e^{-t_i}\cdot \sigma(t_i)^{2},\\
    \label{eq:d3}
    &\frac{\partial H}{\partial o_j} =  \sum_{i=1}^{M}\sum_{j=1}^{N} p_{ij}\cdot \tau_j \cdot \Phi(\theta_i, \delta_j)^{\tau_j \cdot w_j} \cdot \ln(\Phi(\theta_i, \delta_j)) \cdot \hat{p}_{ij} \cdot \sigma(o_j),\\
    \label{eq:d4}
    &\frac{\partial H}{\partial b_j} =  \sum_{i=1}^{M}\sum_{j=1}^{N} p_{ij}\cdot w_j \cdot \Phi(\theta_i, \delta_j)^{\tau_j \cdot w_j} \cdot \ln(\Phi(\theta_i, \delta_j)) \cdot \hat{p}_{ij} \cdot [1 - tanh(b_j)^{2}],
\end{align}

\noindent where: 
\begin{align*}
\refstepcounter{equation}\latexlabel{eq:r1}
\refstepcounter{equation}\latexlabel{eq:r2}
\refstepcounter{equation}\latexlabel{eq:r3}
\Phi(\theta_i, \delta_j) &= \bigg(\frac{\delta_j}{1 - \delta_j}\bigg)\cdot \bigg(\frac{\theta_i}{1 - \theta_i}\bigg)^{-1},~\Theta(\theta_i) = \frac{1}{\theta_i\cdot (1 - \theta_i)},~\Delta(\delta_j)  = \frac{1}{\delta_j \cdot (1 - \delta_j)}.
\tag{\ref*{eq:r1}, \ref*{eq:r2}, \ref*{eq:r3}}
\end{align*}
Given the partial derivatives, the model parameters are updated using gradient descent, according to Equations (\ref{eq:d5}), (\ref{eq:d6}), (\ref{eq:d7}) and (\ref{eq:d8}): 
\begin{align*}
\refstepcounter{equation}\latexlabel{eq:d5}
\refstepcounter{equation}\latexlabel{eq:d6}
\refstepcounter{equation}\latexlabel{eq:d7}
\refstepcounter{equation}\latexlabel{eq:d8}
d_j^{(n+1)} = d_j^{(n)} - \eta\cdot\frac{\partial H}{\partial d_j^{(n)}}, &\qquad t_i^{(n+1)} = t_i^{(n)} - \eta\cdot\frac{\partial H}{\partial t_i^{(n)}},
\tag{\ref*{eq:d5}, \ref*{eq:d6}}\\
o_j^{(n+1)} = o_j^{(n)} - \eta\cdot\frac{\partial H}{\partial o_j^{(n)}}, &\qquad b_j^{(n+1)} = b_j^{(n)} - \eta\cdot\frac{\partial H}{\partial b_j^{(n)}}.
\tag{\ref*{eq:d7}, \ref*{eq:d8}}
\end{align*}

\begin{figure*}[!htb]
\centering
\includegraphics[width=\linewidth,trim={0cm 0.cm 12.cm 0.cm},clip]{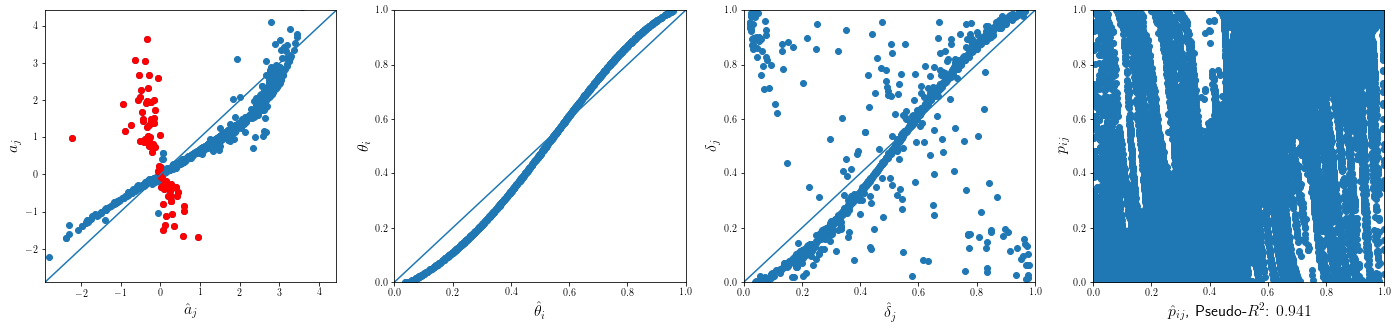}
%
\caption{Scatter plots showing sampled discriminations (left), abilities (centre) and difficulties (right) used to generate a $1000\times1000$ response matrix, and their estimates produced by $\beta^4$-IRT with 1000 initial iterations with fixed discriminations. Red dots on the discrimination plot represent discriminations estimated with flipped signs (77 out of 1000 discriminations). }
\label{fig:b4-1000-init-1000x1000(77).png}
\end{figure*}

Figure \ref{fig:b4-1000-init-1000x1000(77).png} shows that simply refactoring of the discrimination parameter was not enough to solve the symmetry problem as 77 discriminations were wrongly assigned flipped signs. However, the new formulation allows us to select sensible  priors for the parameters that lead to much better estimates:

\begin{itemize}
    \item Abilities: set $t_i^{(0)}$ such that $\theta_i^{(0)} = \sigma(t_i^{(0)}) = N^{-1}\left(\sum_{j=1}^N p_{ij}\right)$;
    
    \item Difficulties: set $d_j^{(0)}$ such that $\delta_j^{(0)} = \sigma(d_j^{(0)}) = 1 - M^{-1}\left(\sum_{i=1}^M p_{ij}\right)$;
    \item Discrimination magnitudes: set $o_j^{(0)}$ such that $\omega_j^{(0)} = \text{softplus}(o_j^{(0)}) = 1$;
    \item Discrimination signs: set $\tau_j = \rho(\vec{\theta}^{(0)}, \vec{p}_j)$, where $\rho$ is the Pearson correlation coefficient, $\vec{\theta}^{(0)} = (\theta_1^{(0)},\ldots, \theta_N^{(0)})$ and $\vec{p}_j = (p_{1j},\ldots, p_{Nj})$.
\end{itemize}

The priors for abilities and difficulties are intuitive as higher abilities lead to higher average responses and the opposite is true for difficulties. As for the discrimination sign parameter, we had two desiderata: (i) they need to capture the fact that for a positive discrimination, expected response grows with ability, while for negative discriminations, higher abilities lead to lower responses; and (ii) their support needs to be in $[-1, 1]$, thus the correlation between abilities and responses for each item lends itself nicely. To avoid flipping the signs during the fitting iterations, especially for very small discriminations that are close to $0$, we keep the $\tau_j$ estimates fixed and only optimise the magnitudes of the discriminations. 

Algorithm \ref{alg:birt-gd} shows the steps of the parameter estimation process  for the $\beta^{4}$-IRT model. Note that the algorithm allows for a certain number of initial iterations where the discrimination parameters are kept fixed, to tackle the symmetry problem and avoid the estimation of discriminations with inverted signs, which can have a negative impact in the estimation of the corresponding difficulties. 

\begin{algorithm}[h]
\caption{$\beta^{4}$-IRT whith \textbf{priors}} 
\label{alg:birt-gd}
\begin{algorithmic}[1]
\Require Response matrix $\mathbf{P}$, where each element $p_{ij}$ corresponds to the observed response from respondent $i$ to item $j$, learning rate $\eta$, number of training epochs with discrimination parameters fixed (\textit{n\_inits}), total number of training epochs (\textit{n\_epochs}).
\State Randomly initialise $t_i^{(0)}$, $d_j^{(0)}$ from $N(0, 1)$;
\State Set $o_j^{(0)}$ such that $\omega_j^{(0)} = $ softplus($o_j^{(0)}$) = 1
\State Set $\tau_j = \rho(\vec{\theta}^{(0)}, \vec{p_j})$ \Comment{where $\rho$ is the Pearson correlation coefficient, $\vec{\theta}^{(0)} = (\theta_1^{(0)},\ldots, \theta_N^{(0)})$ and $\vec{p}_j = (p_{1j},\ldots, p_{Nj})$.}
 
\State set $t_i^{(0)}$ such that $\theta_i^{(0)} = \sigma(t_i^{(0)}) = N^{-1}\left(\sum_{j=1}^N p_{ij}\right)$.
    
\State Set $d_j^{(0)}$ such that $\delta_j^{(0)} = \sigma(d_j^{(0)}) = 1 - M^{-1}\left(\sum_{i=1}^M p_{ij}\right)$.

\State Set $n \leftarrow 0$
\For { $n < $ \textit{n\_epochs} }


    \State Calculate estimated responses using Equation (\ref{eq:fo});
    \State Calculate the loss using Equation (\ref{eq:crossentropy});
    \State Calculate the partial derivatives using Equations Equations (\ref{eq:d1}), (\ref{eq:d2}), (\ref{eq:d3}) and (\ref{eq:d4}); 
    \State Update $d_j^{(n+1)}$ and $t_i^{(n+1)}$ according to Equations (\ref{eq:d5}) and (\ref{eq:d6});
    \If{$n \geq $ \textit{n\_inits}}
    \State Update $o_j^{(n+1)}$ and $b_j^{(n+1)}$ according to Equations (\ref{eq:d7}) and (\ref{eq:d8});
    
    \Else
        \State Set $o_j^{(n+1)}\leftarrow o_j^{(n)}$ and $b_j^{(n+1)}\leftarrow b_j^{(n)}$;
    \EndIf

    \State $n \leftarrow n + 1$;
\EndFor
\State $\theta_i \leftarrow \sigma$($t_i$);
\State $\delta_j \leftarrow \sigma$($d_j$);
\State $a_j \leftarrow \omega_j \cdot \tau_j$;\\
\Return Estimated parameters $\theta_i$, $\delta_j$ and $a_j$.
\end{algorithmic}
\end{algorithm}

\begin{figure*}[!htb]
\centering
\includegraphics[width=\linewidth,trim={0cm 0.cm 12.cm 0.cm},clip]{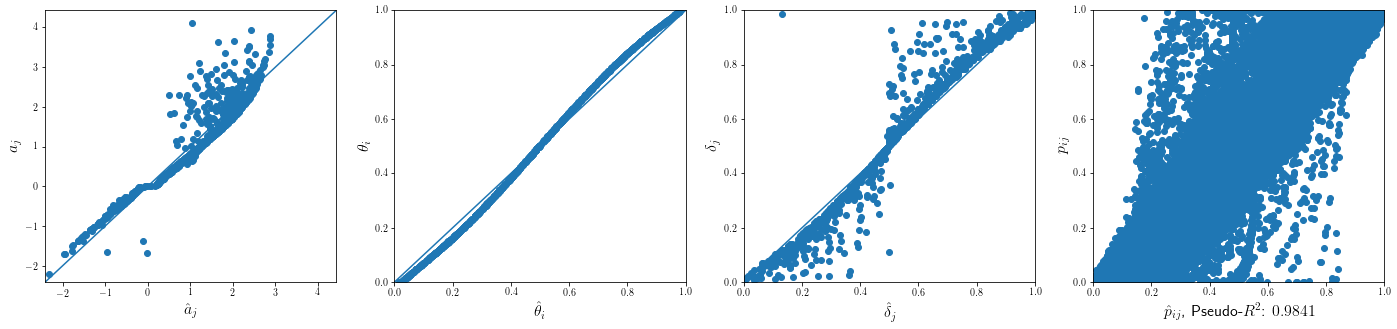}
%
\caption{Scatter plots showing sampled discriminations (left), abilities (centre) and difficulties (right) used to generate a $1000\times1000$ response matrix, and their estimates produced by $\beta^4$-IRT with 1000 initial iterations with fixed discriminations and better priors for all parameters. All discriminations were estimated with the correct signs. }
\label{fig:b4-priors-1000-init-1000x1000(0).png}
\end{figure*}

Figure \ref{fig:b4-priors-1000-init-1000x1000(0).png} shows the resulting estimates after fitting $\beta^4$-IRT with the above priors. No discriminations where estimated with flipped signs, which led to better estimates overall. Due to these good results, from here on in this paper, we refer to this version when we mention $\beta^4$-IRT.

\subsection{Checking goodness of fit}
\label{sec:pseudor2}


 To the best of our knowledge, the previous IRT approaches did not consider an $R^2$ metric to evaluate the goodness-of-fit for a model. The $R^2$ gives an easy interpretation of the model's performance and can be used to compare two or more IRT models. 
 Here, we  propose a Pseudo-$R^{2}$,  given by Equation (\ref{eq:pseudo:r2}) 

    
\begin{equation}\label{eq:pseudo:r2}
    \text{Pseudo-}R^{2} = 1 - \frac{u}{v},
\end{equation}

\noindent where $u$ is the sum of squared residues and $v$ is the sum of the quadratic differences from the mean, defined respectively by Equations (\ref{eq:pseudo:r2:u}) and (\ref{eq:pseudo:r2:v}):

\begin{align*}
\refstepcounter{equation}\latexlabel{eq:pseudo:r2:u}
\refstepcounter{equation}\latexlabel{eq:pseudo:r2:v}
u = \sum_{i=1}^{M}\sum_{j=1}^{N}(p_{ij} - \hat{p}_{ij})^{2} \mbox{ and } & v = \sum_{i=1}^{M}\sum_{j=1}^{N}(p_{ij} - \bar{p})^{2},
\tag{\ref*{eq:pseudo:r2:u}, \ref*{eq:pseudo:r2:v}}
\end{align*}

    

\noindent where $\bar{p}$ is the mean of the observed response. According to \cite{pseudor2}, the Pseudo-$R^{2}$  can be interpreted as the square of the correlation between the estimated ($\hat{p}$) and the observed ($p$) values. The denominator $v$ can be seen as the quadratic error of the null model. 

\subsection{Model implementation in Python}

We implemented $\beta^{4}$-IRT using the automatic differentiation capabilities of the Python library TensorFlow. In this section, we describe the installation process and present a short tutorial on how to fit the model and use the tools provided by the package {\tt birt-gd}. 


The package can be downloaded from Python's Package Index (\url{https://pypi.org/project/birt-gd/}) or by cloning the repository on  GitHub in \url{https://github.com/Manuelfjr/birt-gd}. It is also possible to directly install {\tt birt-gd} using the following command line:

\begin{CodeChunk}
\begin{CodeInput}
# pip install birt-gd
\end{CodeInput}
\end{CodeChunk}

To use the package, first we need to import it in Python. Below we show an example of use:

\begin{CodeChunk}
\begin{CodeInput}
>>> from birt import Beta4
>>> data = pd.DataFrame({'a': [0.99,0.89,0.87],
...                      'b': [0.32,0.25,0.45]})
>>> b4 = Beta4(n_models = 2, 
...            n_instances = 3, 
...            random_seed=1)
>>> b4.fit(data.values)
\end{CodeInput}
\begin{CodeOutput}
2\%||       | 119/5000 [00:01<01:05, 74.58it/s]Model converged at the 122th epoch

2\%||       | 122/5000 [00:01<01:07, 72.35it/s]
<birt.Beta4 object at 0x7f420baa3b50>
\end{CodeOutput}
\begin{CodeInput}
>>> b4.abilities
\end{CodeInput}
\begin{CodeOutput}
array([0.8940176, 0.2747254], dtype=float32)
\end{CodeOutput}
\begin{CodeInput}
>>> b4.difficulties
\end{CodeInput}
\begin{CodeOutput}
array([0.38353133, 0.5238179 , 0.37623164], dtype=float32)
\end{CodeOutput}
\begin{CodeInput}
>>> b4.discriminations
\end{CodeInput}
\begin{CodeOutput}
array([1., 1., 1.], dtype=float32)
\end{CodeOutput}
\end{CodeChunk}

We now illustrate an example with more data, to better explain the module's features. First, we create $5$ respondents and $20$ items by randomly sampling their abilities and difficulties from Beta distributions. For the abilities, we sample the first one from $\Beta(1, 0.1)$, the second from $\Beta(1, 10)$ and the remaining three abilities from $\Beta(1, 1)$. For the difficulties, we randomly sample the first one from $\Beta(1, 10)$, the second from $\Beta(1, 5)$ and the remaining ones from $\Beta(1, 1)$. These values were chosen such that respondent $i=0$ will likely have high ability and item $j=0$ will likely have low difficulty. Finally, we sample the 20 items' discriminations from $\mathcal{N}(1, 1)$.

\begin{CodeChunk}
\begin{CodeInput}
>>> import numpy as np
>>> import pandas as pd
>>> from birt import BIRTGD
>>> import matplotlib.pyplot as plt
>>> m, n = 5, 20
>>> np.random.seed(1)
>>> abilities = [np.random.beta(1, i) for i in ([0.1, 10] + [1] * (m - 2))]
>>> difficulties = [np.random.beta(1, i) for i in [10, 5] + [1] * (n - 2)]
>>> discriminations = list(np.random.normal(1, 1, size = n))
\end{CodeInput}
\end{CodeChunk}

Then we calculate the expected responses of the $5$ respondents for the $20$ items, using Equation (\ref{eq:ep_birt}), yielding response matrix $\mathbf{P}_{5 \times 20}$ (called \textsf{pij} in the code), where each value is the observed response of the $i$-th respondent for the $j$-th item.

\begin{CodeChunk}
\begin{CodeInput}
import numpy as np

m, n = 5, 20
np.random.seed(1)
abilities = [np.random.beta(1, i) for i in ([0.1, 10] + [1] * ( m - 2))]
difficulties = [np.random.beta(1, i) for i in [10, 5] + [1] * (n - 2)]
discrimination = list(np.random.normal(1, 1, size = n))
pij = pd.DataFrame(columns = range( m ), index = range( n ))

>>> i, j = 0, 0
>>> for theta in abilities:
>>>   for delta, a in zip(difficulties, discrimination):
>>>     alphaij = ( theta/delta )** (a)
>>>     betaij = ((1 - theta)/(1 - delta)) ** (a)
>>>     pij.loc[j, i] =  np.mean(np.random.beta(alpha, beta, size = 100) )[0]
>>>     j+=1
>>>   j = 0
>>>   i+=1
\end{CodeInput}
\end{CodeChunk}

We then use class \textsf{BIRTGD} from the \textsf{birt} module to fit a model on the observed responses, with learning rate $\eta=1$, $5000$ total training epochs and $1000$ initial epochs with fixed discrimination parameters. Note that the default values for the arguments \textsf{epochs} and \textsf{n\_inits} are 10000 and 1000, respectively.

\begin{CodeChunk}
\begin{CodeInput}
>>> b4 = Beta4(
...        learning_rate = 1, 
...        epochs = 5000,
...        n_respondents = pij.shape[1], 
...        n_items = pij.shape[0],
...        n_inits = 1000, 
...        n_workers = -1,
...        random_seed = 1,
...        tol= 10 ** (-8),
...        set_priors = False
...        )
>>> b4.fit( pij )

\end{CodeInput}
\end{CodeChunk}

After fitting the model we can check the \textsf{score} attribute, which returns the corresponding Pseudo-$R^{2}$, as discussed in section \ref{sec:pseudor2}.

\begin{CodeChunk}
\begin{CodeInput}
>>> b4.score
\end{CodeInput}
\begin{CodeOutput}
0.9038146230196351
\end{CodeOutput}
\end{CodeChunk}

In this case, the model fit this small dataset very well, with Pseudo-$R^{2} > 0.9$. We can also view some descriptive statistics using the \textsf{summary} method, in similar fashion to R's \textsf{summary} function, including the the Pseudo-$R^{2}$ value and the quartiles, minima and maxima of the estimated abilities, difficulties, discriminations and responses. 

\begin{CodeChunk}
\begin{CodeInput}
>>> b4.summary()
\end{CodeInput}
\begin{CodeOutput}
        ESTIMATES
        -----
                        | Min      1Qt      Median   3Qt      Max      Std.Dev
        Ability         | 0.00010  0.22147  0.63389  0.73353  0.92040  0.33960
        Difficulty      | 0.01745  0.28047  0.63058  0.84190  0.98624  0.31635
        Discrimination  | 0.31464  1.28330  1.61493  2.22936  4.44645  1.02678
        pij             | 0.00000  0.02219  0.35941  0.86255  0.99993  0.40210
        -----
        Pseudo-R2       | 0.90381
\end{CodeOutput}
\end{CodeChunk}



From the \textsf{summary} output above, we note that the statistics of the estimated abilities and difficulties were close to those of a $\Beta(1, 1)$, which was the distribution from which most of the simulated values for these parameters were sampled, with a shift in the median (which for a $\Beta(1, 1)$ is equal to $0.5$), because two abilities were sampled from $\Beta(1, 0.1)$ and $\Beta(1, 10)$ and two difficulties were sampled from $\Beta(1, 10)$ and $\Beta(1, 5)$.

\begin{figure}[h!]
    \centering
    \includegraphics[scale=0.5]{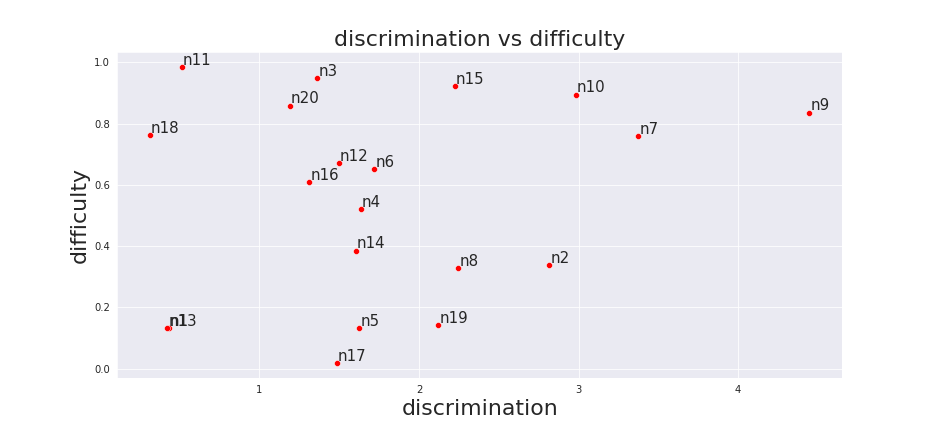}
    \caption{Estimated discrimination and difficulty values for each item.}
    \label{fig:disc_vs_diff}
\end{figure}

In addition to descriptive information, the module provides functions to create some useful plots to help analyse each parameter. The code chunks below show examples of these plots. 
First, we show how to create a scatter plot of the estimated item discriminations ($x$ axis) and difficulties ($y$ axis). The resulting plot in Figure \ref{fig:disc_vs_diff} shows an apparently uncorrelated distribution, without negative discriminations and with the presence of a possible discrimination outlier. 

\begin{CodeChunk}
\begin{CodeInput}
>>> import matplotlib.pyplot as plt
>>> b4.plot(xaxis = 'discrimination',
...           yaxis = 'difficulty',
...           ann = True,
...           kwargs = {'color': 'red'},
...           font_size = 22, font_ann_size = 15)
>>> plt.show()
\end{CodeInput}
\end{CodeChunk}

\begin{figure}[h!]
    \centering
    \includegraphics[scale=0.5]{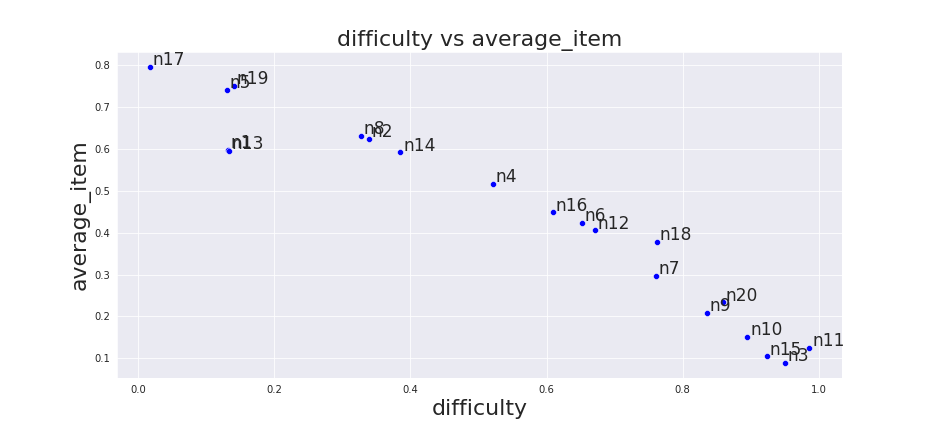}
    \caption{Estimated difficulty values and average response for each item.}
    \label{fig:diff_vs_average}
\end{figure}

The next example shows how to draw the scatter plot shown by
 Figure \ref{fig:diff_vs_average}, where a strong negative linear relationship can be seen between difficulty and the average response for each item.

\begin{CodeChunk}
\begin{CodeInput}
>>> b4.plot(xaxis = 'difficulty', yaxis = 'average_item',
...           ann = True, kwargs = {'color': 'blue'},
...           font_size = 22, font_ann_size = 17)
>>> plt.show()
\end{CodeInput}
\end{CodeChunk}

\begin{figure}[h!]
    \centering
    \includegraphics[scale=0.5]{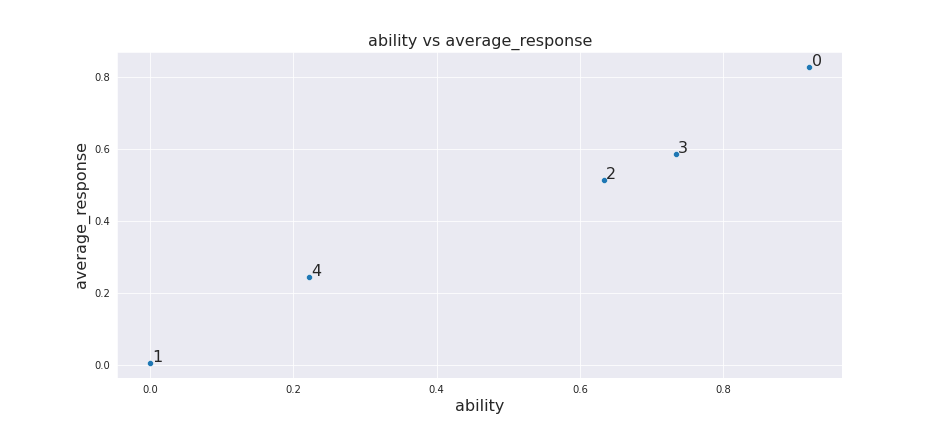}
    \caption{Estimated ability values and average response for each respondent.}
    \label{fig:ab_vs_average}
\end{figure}

According to Figure \ref{fig:ab_vs_average} (see the code below), we observe a strong positive linear relationship between the respondent ability and the average response. 

\begin{CodeChunk}
\begin{CodeInput}
>>> b4.plot(xaxis = 'ability', yaxis = 'average_response',
...           ann = True, font_size = 16, font_ann_size = 16)
>>> plt.show()
\end{CodeInput}
\end{CodeChunk}

For the scatter plots, the arguments \textbf{xaxis}
and \textbf{yaxis} define the variable that will occupy the $x$ and $y$ axes in the graphic, respectively. Argument \textbf{ann} is a boolean value used to define if the graph's points should be plotted alongside their indexes in the data set. 
Finally, \textbf{kwargs} is a dictionary with keyword arguments, which is familiar for Matplotlib \citep{Hunter:2007} users, and can be used to pass any keyword arguments that can be used by Matplotlib. In addition to scatter plots, we can plot boxplots for the estimated abilities, difficulties and discrminations.

\begin{CodeChunk}
\begin{CodeInput}
>>> b4.boxplot(y = 'ability',
...              kwargs = {'linewidth': 4}, font_size = 27)
>>> b4.boxplot(x = 'difficulty', font_size = 27)
>>> b4.boxplot(y = 'discrimination', font_size = 27)
\end{CodeInput}
\end{CodeChunk}

\begin{figure}[htp]
    \centering
    \begin{subfigure}[b]{0.45\linewidth}
        \includegraphics[width=\textwidth]{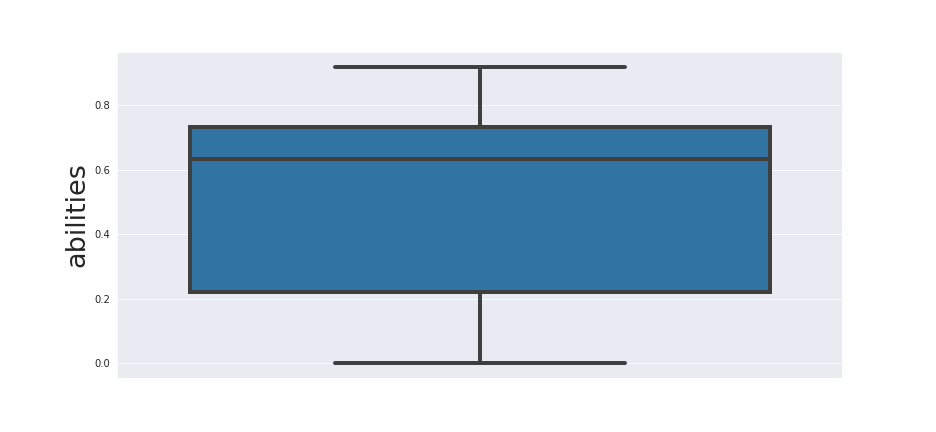}
        \caption{Boxplot of the estimated abilities}
        \label{fig:boxplot_ab}
    \end{subfigure}
    \hfill
    \begin{subfigure}[b]{0.45\linewidth}
        \includegraphics[width=\textwidth]{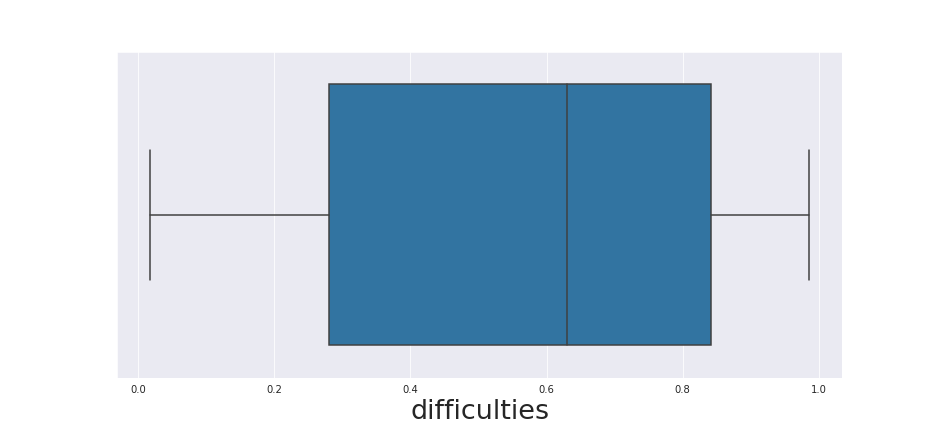}
        \caption{Boxplot of the estimated  difficulties}
        \label{fig:boxplot_diff}
    \end{subfigure}
    \vfill
    \begin{subfigure}[b]{0.45\linewidth}
        \includegraphics[width=\textwidth]{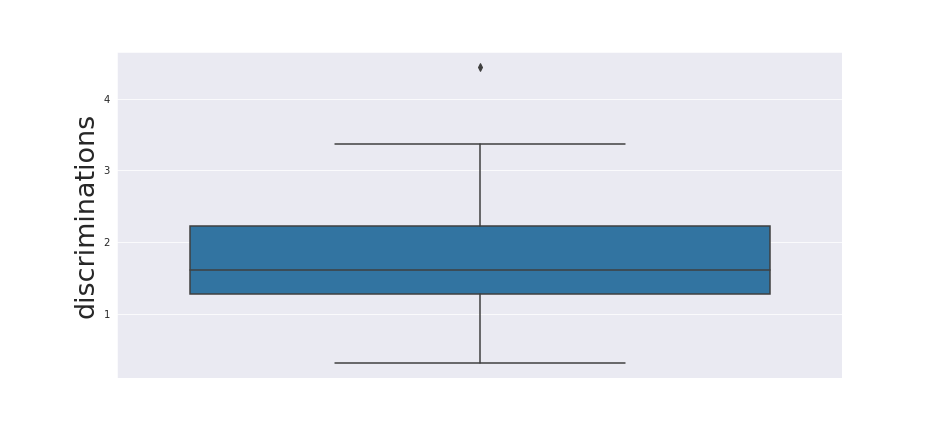}
        \caption{Boxplot of the estimated  discrimination}
        \label{fig:boxplot_disc}
    \end{subfigure}
    \caption{Boxplots of the estimates for the $\beta^{4}$-IRT parameters.}
    \label{fig:boxplots}
\end{figure}




 As in scatter plots, boxplots also have the $x$ and $y$ arguments, as well as the \textbf{kwargs} dictionary.

\section{Evaluating parameter recovery}\label{sec:experiments}

In this Section we assess the performances of $\beta^{3}-$BIRT and $\beta^{4}-$BIRT in recovering the actual item and respondent parameters. For $\beta^{3}-$BIRT we use the version provided in our {\tt birt-gd} package, which user a number of initialisation iterations with fixed discriminations, as mentioned in Section \ref{sec:irt}.


A Monte Carlo experiment with $30$ replications was employed to evaluate the performances of the four models taking into account three dataset configurations:  $N=100$ items and $M=20$ respondents (dataset 1), $N=100$ items and $M=100$ respondents (dataset 2),  and finally, $N=300$ items and $M=50$ respondents (dataset 3). 


For each dataset and Monte Carlo replication, we sample respondent abilities and item difficulties from $\Beta(1, 1)$ and item discriminations from $\mathcal{N}(1, 1)$. Then, for each response $p_{ij}$, we take the mean of $100$ samples taken from beta distributions as described in Equation (\ref{eq:model_def}). The resulting response matrix $\mathbf{P}$ is used to fit $\beta^{3}-$IRT and $\beta^{4}-$BIRT,  using $50000$ epochs and $1000$ initialisations with fixed discriminations. All the other hyperparameters were set as their default values.  


Using bootstrap \citep{hesterberg2011bootstrap}, we calculated the 95\% confidence interval for the Pearson correlation $\rho$ between estimated and original parameter values. The aim is to measure how well the models recover the original parameter rankings. 
In addition, we also considered a 95\% confidence interval for the Relative Squared Error (RSE) to evaluate the quality of the parameter estimates for each model. The RSE represents the proportion of the unexplained variance, being defined by $RSE = 1 - R^2$. 

\begin{table}[!htb]
    \centering
    
    \begin{tabular}{ccccc}
\hline
Dataset & Parameter & Model &RSE, 95\% CI&  $\rho$, 95\% CI\\
\hline
\multirow{6}{*}{N = 100, M = 100} & \multirow{2}{*}{$a_j$} & $\beta^{3}$-IRT &  [0.2295, 0.3136] &  \textbf{[0.9697, 0.9768]} \\
                &         & $\beta^{4}$-IRT &  \textbf{[0.0999, 0.1305]} &  [0.9496, 0.9636] \\
\cline{2-5}
                & \multirow{2}{*}{$\delta_j$} & $\beta^{3}$-IRT &  [0.3971, 0.5393] &   [0.7019, 0.7860] \\
                &         & $\beta^{4}$-IRT &  \textbf{[0.0947, 0.1125] }&  \textbf{[0.9738, 0.9795]} \\
\cline{2-5}
                & \multirow{2}{*}{$\theta_i$} & $\beta^{3}$-IRT &  \textbf{[0.0379, 0.0531]} &   \textbf{[0.9957, 0.9970]} \\
                &         & $\beta^{4}$-IRT &   [0.0670, 0.0805] &   [0.9918, 0.9930] \\
\cline{1-5}
\cline{2-5}
\multirow{6}{*}{N = 100, M = 20} & \multirow{2}{*}{$a_j$} & $\beta^{3}$-IRT &    [0.1200, 0.1831] &  \textbf{[0.9661, 0.9767] }\\
                &         & $\beta^{4}$-IRT &   \textbf{[0.1220, 0.1623]} &    [0.9500, 0.9622] \\
\cline{2-5}
                & \multirow{2}{*}{$\delta_j$} & $\beta^{3}$-IRT &  [0.3401, 0.4646] &  [0.7494, 0.8207] \\
                &         & $\beta^{4}$-IRT &  \textbf{[0.1045, 0.1307] }& \textbf{ [0.9678, 0.9754]} \\
\cline{2-5}
                & \multirow{2}{*}{$\theta_i$} & $\beta^{3}$-IRT & \textbf{ [0.0389, 0.0702]} &  \textbf{[0.9949, 0.9974]} \\
                &         & $\beta^{4}$-IRT &  [0.0865, 0.1193] &  [0.9898, 0.9937] \\
\cline{1-5}
\cline{2-5}
\multirow{6}{*}{N = 300, M = 50} & \multirow{2}{*}{$a_j$} & $\beta^{3}$-IRT &  \textbf{[0.1038, 0.1295] }& \textbf{ [0.9666, 0.9734] }\\
                &         & $\beta^{4}$-IRT &   [0.1380, 0.1602] &   [0.9534, 0.9620] \\
\cline{2-5}
                & \multirow{2}{*}{$\delta_j$} & $\beta^{3}$-IRT &  [0.3107, 0.3923] &  [0.7859, 0.8339] \\
                &         & $\beta^{4}$-IRT &  \textbf{[0.1029, 0.1206]} &  \textbf{[0.9675, 0.9726]} \\
\cline{2-5}
                & \multirow{2}{*}{$\theta_i$} & $\beta^{3}$-IRT & \textbf{ [0.0168, 0.0284] }&  \textbf{[0.9979, 0.9988]} \\
                &         & $\beta^{4}$-IRT &  [0.0656, 0.0863] &  [0.9916, 0.9937] \\
\hline
\end{tabular}

    \caption{95\% confidence intervals for RSE and $\rho$ calculated using bootstrap. The best model is marked in \textbf{bold}.}
\label{tab:rse:rho}
\end{table}

Table \ref{tab:rse:rho} shows our results. Its is clear that, for the difficulty parameter, $\beta^{4}$-IRT outperformed $\beta^{3}$-IRT in all cases, while $\beta^{3}$-IRT was better at estimating abilities, according to RSE and $\rho$. For discriminations, while there no overall best-performing method, Table \ref{tab:sign} shows that, as expected, $\beta^{4}$-IRT was much better than $\beta^{3}$-IRT at correctly predicting the signs of the discrimination parameters, never switching more than $0.06\%$ of the signs, which given the number of items in these datasets (100, 100 and 300), means that most runs estimated all signs correctly. This had a very significant impact on the estimation of difficulties, which showed the highest difference between the confidence intervals of both methods in Table \ref{tab:rse:rho}.

\begin{table}[!htb]
\centering
    \begin{tabular}{ccc}
\hline
Dataset & Model &  Inverted signs (\%)  \\
\hline
\multirow{2}{*}{N = 100, M = 100} & $\beta^{3}$-IRT &  [3.2333, 4.7667] \\
                & $\beta^{4}$-IRT&  \textbf{[0.0333, 0.2667]}  \\
\cline{1-3}
\multirow{2}{*}{N = 100, M = 20} & $\beta^{3}$-IRT &     [3.0333, 4.400]  \\
                & $\beta^{4}$-IRT&     \textbf{[0.1000, 0.5333] } \\
\cline{1-3}
\multirow{2}{*}{N = 300, M = 50} & $\beta^{3}$-IRT &  [2.5333, 3.4889] \\
                & $\beta^{4}$-IRT&  \textbf{[0.0556, 0.2222]}  \\
\hline
\end{tabular}

    \caption{95\% confidence intervals for proportion of discriminations estimated with inverted signs.}
\label{tab:sign}
\end{table}

\section{Summary and discussion} \label{sec:summary}

$\beta^{4}$-IRT and $\beta^{3}$-IRT implemented in Python, resulting in a package that was published in the official repository\footnote{\url{https://pypi.org/project/birt-gd/}} of the languague. Experiments showed that, although $\beta^{3}$-IRT and $\beta^{4}$-IRT performed similarly when estimating abilities and discrimination values, $\beta^{4}$-IRT presented a superior performance in the recovery of discrimination signs, which led to an improvement in difficulty estimation, according to RSE and $\rho$. Improving the estimation of discrimination signs can be important for certain applications of IRT. For example, \cite{chen2019birt} investigated the interpretation of negatively-discriminated items as noisy instances in a dataset, i.e. instances that might have flipped labels, making their classification harder for the best models in a model pool. This analysis is clearly hindered if the IRT model is unable to correctly identify the signs of the items' discriminations. 


\section*{Computational details}


The results of this paper were obtained using Python $3.6$, but can be reproduced in any version higher than $3.6$. The module has the following dependencies: {Numpy} ($\geq1.19.5$), {tqdm} ($\geq1.19.5$), {tensorflow} ($\geq4.59.0$), {pandas} ($\geq1.2.3$), {seaborn} ($\geq0.11.0$), {matplotlib} ($\geq3.3.2$) and {scikit-learn} ($\geq0.23.2$). All libraries used are available in the Python Package Index (PyPi) at \url{https://pypi.org/}.


\section*{Acknowledgments}


MFJ would like to thank the Brazilian National Council for Scientific and Technological Development (CNPq) for their financial support through grant number \textit{PIA12073-2020}.




\bibliographystyle{unsrtnat}
\bibliography{refs}


\newpage

\begin{appendix}
\section{Appendix} \label{app:appendix}

About the  (\ref{eq:crossentropy}) equation, we need to calculate the partial derivates with respect to $t_i$, $d_j$, $b_j$ e $o_j$. Then, we have:

\begin{proof}[For $t_i$:]
    \begin{equation}
            \label{eq:ap1}
            \frac{\partial H}{\partial t_i} = -\sum_{i=1}^{M}\sum_{j=1}^{N} p_{ij}\cdot \frac{1}{E[\hat{p}_{ij} | t_i, d_j, o_j, b_j]} \cdot \frac{\partial E[\hat{p}_{ij} | t_i, d_j, o_j, b_j]}{\partial t_i};
    \end{equation}
        
    such that $E = E[\hat{p}_{ij} | t_i, d_j, o_j, b_j]$, 
    
        \begin{equation}
            \frac{\partial E}{\partial t_i} = \frac{0 - ( - \tau_j w_j)\cdot\big(\frac{\theta_i}{1 - \theta_i}\big)^{-\tau_j w_j - 1}\cdot \bigg[\frac{\partial \theta_i}{\partial t_i}\cdot (1 - \theta_i) - ( -1)\cdot \frac{\partial \theta_i}{\partial t_i}\cdot (\theta_i)\bigg]\cdot \big(\frac{1}{1 - \theta_i}\big)^{2}\cdot\big(\frac{\delta_j}{1 - \delta_j}\big)^{\tau_j w_j}}{
            \bigg[1 + \big(\frac{\delta_j}{1 - \delta_j}\big)^{\tau_j w_j}\cdot \big(\frac{\theta_j}{1 - \theta_j}\big)^{-\tau_j w_j}\bigg]^{2}
            } = 
        \end{equation}
        \begin{equation}
            = \frac{\tau_j w_j\cdot \big(\frac{\delta_j}{1 - \delta_j}\big)^{\tau_j w_j}\cdot \big(\frac{\theta_i}{1 - \theta_i}\big)^{-\tau_j w_j} \cdot \big(\frac{1 - \theta_i}{\theta_i}\big) \cdot \big(\frac{1}{1 - \theta_i}\big)^{2}\cdot \big[\frac{\partial \theta_i}{\partial t_i}\big]}{
            \bigg[1 + \big(\frac{\delta_j}{1 - \delta_j}\big)^{\tau_j w_j}\cdot \big(\frac{\theta_j}{1 - \theta_j}\big)^{-\tau_j w_j}\bigg]^{2}
            } = 
        \end{equation}
        
        \begin{equation}
        \label{eq:f1}
            = \frac{\tau_j w_j\cdot \bigg[\big(\frac{\delta_j}{1 - \delta_j}\big)\cdot \big(\frac{\theta_i}{1 - \theta_i}\big)^{-1}\bigg]^{\tau_j w_j}\cdot  \big[\frac{1}{\theta_i\cdot(1 - \theta_i)}\big]\cdot \big[\frac{\partial \theta_i}{\partial t_i}\big]}{
            \bigg[1 + \big(\frac{\delta_j}{1 - \delta_j}\big)^{\tau_j w_j}\cdot \big(\frac{\theta_j}{1 - \theta_j}\big)^{-\tau_j w_j}\bigg]^{2}
            } = 
        \end{equation}
    
    Note that:
    \begin{equation}
        E^{2} = \bigg[\frac{1}{1 + \big(\frac{\delta_j}{1 - \delta_j}\big)^{\tau_j w_j}\cdot \big(\frac{\theta_i}{1 - \theta_i}\big)^{-\tau_j w_j}}\bigg]^{2}
    \end{equation}
    
    Furthermore, replacing the equations (\ref{eq:r1}) and (\ref{eq:r2}) in (\ref{eq:f1}), then:
    
    \begin{equation}
        \label{eq:e_ti}
        \frac{\partial E}{\partial t_i}  = \tau_j w_j \cdot \Phi(\theta_i, \delta_j)^{\tau_j w_j} \cdot \Theta(\theta_i) \cdot E^{2} \cdot \frac{\partial \theta_i}{\partial t_i}
    \end{equation}
    
    So, replacing (\ref{eq:e_ti}) in (\ref{eq:ap1}), we have:
    \begin{equation}
        \frac{\partial H}{\partial t_i} = - \sum_{i=1}^{M}\sum_{j=1}^{N} p_{ij}\cdot \frac{1}{E[\hat{p}_{ij} | t_i, d_j, o_j, b_j]} \cdot \tau_j w_j \cdot \Phi(\theta_i, \delta_j)^{\tau_j w_j} \cdot \Theta(\theta_i) \cdot E[\hat{p}_{ij} | t_i, d_j, o_j, b_j]^{2} \cdot \frac{\partial \theta_i}{\partial t_i} = 
    \end{equation}
    \begin{equation}
         = -\sum_{i=1}^{M}\sum_{j=1}^{N} p_{ij}\cdot \tau_j w_j \cdot \Phi(\theta_i, \delta_j)^{\tau_j w_j} \cdot \Theta(\theta_i) \cdot E[\hat{p}_{ij} | t_i, d_j, o_j, b_j]\cdot \frac{\partial \theta_i}{\partial t_i} 
    \end{equation}
    
    Then, we calculate $\frac{\partial \theta_i}{\partial t_i}$ as:
    
    \begin{equation}
        \frac{\partial \theta_i}{\partial t_i} = \frac{0 - (-1)\cdot e^{-t_i}}{(1 + e^{-t_i})^{2}} = \frac{e^{t_i}}{(1 + e^{-t_i})^{2}} = e^{-t_i}\cdot\sigma(t_i)^{2}
    \end{equation}
\end{proof}

\begin{proof}[For $d_j$:]

    \begin{equation}
        \label{eq:ap2}
            \frac{\partial H}{\partial d_i} = - \sum_{i=1}^{M}\sum_{j=1}^{N} p_{ij}\cdot \frac{1}{E[\hat{p}_{ij} | t_i, d_j, o_j, b_j]} \cdot \frac{\partial E[\hat{p}_{ij} | t_i, d_j, o_j, b_j]}{\partial d_i};
    \end{equation}

    We can replicate the last steps for $t_i$, so:

    \begin{equation}
        \frac{\partial E}{\partial d_j} = \frac{0 - (\tau_j w_j)\cdot\big(\frac{\delta_j}{1 - \delta_j}\big)^{\tau_j w_j - 1}\cdot \bigg[\frac{\partial \delta_j}{\partial d_i}\cdot (1 - \delta_j) - ( -1)\cdot \frac{\partial \delta_j}{\partial d_i}\cdot (\delta_j)\bigg]\cdot \big(\frac{1}{1 - \delta_j}\big)^{2}\cdot\big(\frac{\theta_i}{1 - \theta_i}\big)^{-\tau_j w_j}}{
        \bigg[1 + \big(\frac{\delta_j}{1 - \delta_j}\big)^{\tau_j w_j}\cdot \big(\frac{\theta_j}{1 - \theta_j}\big)^{-\tau_j w_j}\bigg]^{2}
            } = \dots = 
    \end{equation}
    \begin{equation}
        \label{eq:f2}
        = -\frac{\tau_j w_j\cdot \bigg[\big(\frac{\delta_j}{1 - \delta_j}\big)\cdot \big(\frac{\theta_i}{1 - \theta_i}\big)^{-1}\bigg]^{\tau_j w_j}\cdot  \big[\frac{1}{\delta_j\cdot(1 - \delta_j)}\big]\cdot \big[\frac{\partial \delta_j}{\partial d_i}\big]}{
            \bigg[1 + \big(\frac{\delta_j}{1 - \delta_j}\big)^{\tau_j w_j}\cdot \big(\frac{\theta_j}{1 - \theta_j}\big)^{-\tau_j w_j}\bigg]^{2}
            } = 
    \end{equation}

    Replacing the (\ref{eq:r1}) and (\ref{eq:r3}) equations in (\ref{eq:f2}), we have:

    \begin{equation}
        \label{eq:e_dj}
        \frac{\partial E}{\partial d_j}  = -\tau_j w_j \cdot \Phi(\theta_i, \delta_j)^{\tau_j w_j} \cdot \Delta(\delta_j) \cdot E^{2} \cdot \frac{\partial \delta_j}{\partial d_j}
    \end{equation}    

    Then, replacing (\ref{eq:e_dj}) in (\ref{eq:ap2}), we have:

    \begin{equation}
        \frac{\partial H}{\partial d_j} = - \sum_{i=1}^{M}\sum_{j=1}^{N} - p_{ij}\cdot \tau_j w_j \cdot \Phi(\theta_i, \delta_j)^{\tau_j w_j} \cdot \Delta(\delta_j) \cdot E[\hat{p}_{ij} | t_i, d_j, o_j, b_j]\cdot \frac{\partial \delta_j}{\partial d_j}
    \end{equation}

    Finally, we calculate $\frac{\partial \delta_j}{\partial d_j}$ as:

    \begin{equation}
            \frac{\partial \delta_j}{\partial d_j} = \frac{0 - (-1)\cdot e^{-d_j}}{(1 + e^{-d_j})^{2}} = \frac{e^{d_j}}{(1 + e^{-d_j})^{2}} = e^{-d_j}\cdot\sigma(d_j)^{2}
    \end{equation}

\end{proof}

\begin{proof}[Para $o_j$:]
    \begin{equation}
        \label{eq:ap3}
            \frac{\partial H}{\partial o_j} = - \sum_{i=1}^{M}\sum_{j=1}^{N} p_{ij}\cdot \frac{1}{E[\hat{p}_{ij} | t_i, d_j, o_j, b_j]} \cdot \frac{\partial E[\hat{p}_{ij} | t_i, d_j, o_j, b_j]}{\partial o_j};
    \end{equation}
    
    Note that:
    \begin{equation}
        E = E[\hat{p}_{ij} | t_i, d_j, o_j, b_j] = \frac{1}{1 + \bigg[\big(\frac{\delta_j}{1 - \delta_j}\big)\cdot \big(\frac{\theta_i}{1 - \theta_i}\big)^{-1}\bigg]^{\tau_j w_j}}
    \end{equation}
    
    So,
    \begin{equation}
        \frac{\partial E}{\partial o_j} = \frac{0 - \text{ln}\big[\big(\frac{\delta_j}{ 1 - \delta_j}\big)\cdot \big(\frac{\theta_i}{ 1 - \theta_i}\big)^{-1}\big] \cdot \big[\big(\frac{\delta_j}{ 1 - \delta_j}\big)\cdot \big(\frac{\theta_i}{ 1 - \theta_i}\big)^{-1}\big]^{\tau_j w_j} \cdot \tau_j \cdot\frac{\partial w_j}{\partial o_j}}{\big[ 1 + \big(\frac{\delta_j}{1 - \delta_j}\big)^{\tau_j w_j}\cdot\big(\frac{\theta_i}{1 - \theta_i}\big)^{-\tau_j w_j}\big]^{2}}
    \end{equation}
    
    Simplifying, using the (\ref{eq:fo}) equations and (\ref{eq:r1}), we have the expression below:    \begin{equation}
        \label{eq:e_oj}    
        \frac{\partial E}{\partial o_j} = - E^{2}\cdot \tau_j \cdot \text{ln}[\Phi(\theta_i, \delta_j)] \cdot \big[\Phi(\theta_i, \delta_j)]^{\tau_j w_j} \cdot \frac{\partial w_j}{\partial o_j}
    \end{equation}
    
    Then, replacing the equation (\ref{eq:e_oj}) in (\ref{eq:ap3}), we have:

    \begin{equation}
            \frac{\partial H}{\partial o_j} = - \sum_{i=1}^{M}\sum_{j=1}^{N} -p_{ij}\cdot E\cdot \tau_j \cdot \text{ln}[\Phi(\theta_i, \delta_j)] \cdot \big[\Phi(\theta_i, \delta_j)]^{\tau_j w_j} \cdot \frac{\partial w_j}{\partial o_j}
    \end{equation}
    
    Finally, we can calculate $\frac{\partial w_j}{\partial o_j}$ as:
    
    \begin{equation}
        \frac{\partial w_j}{\partial o_j} = \frac{\partial (\text{ln}(1 + e^{o_j}) )}{\partial o_j} = \frac{e^{o_j}}{1 + e^{o_j}} = \frac{1}{1 + e^{-o_j}} = \sigma(o_j)
    \end{equation}
\end{proof}

\begin{proof}[For $b_j$:]
    \begin{equation}
        \label{eq:ap4}
            \frac{\partial H}{\partial b_j} = - \sum_{i=1}^{M}\sum_{j=1}^{N} p_{ij}\cdot \frac{1}{E[\hat{p}_{ij} | t_i, d_j, o_j, b_j]} \cdot \frac{\partial E[\hat{p}_{ij} | t_i, d_j, o_j, b_j]}{\partial b_j};
    \end{equation}
    
    Similarly, as done for $o_j$, we get the next expression:
    
    \begin{equation}
        \frac{\partial E}{\partial b_j} = \frac{0 - \text{ln}\big[\big(\frac{\delta_j}{ 1 - \delta_j}\big)\cdot \big(\frac{\theta_i}{ 1 - \theta_i}\big)^{-1}\big] \cdot \big[\big(\frac{\delta_j}{ 1 - \delta_j}\big)\cdot \big(\frac{\theta_i}{ 1 - \theta_i}\big)^{-1}\big]^{\tau_j w_j} \cdot w_j \cdot\frac{\partial \tau_j}{\partial b_j}}{\big[ 1 + \big(\frac{\delta_j}{1 - \delta_j}\big)^{\tau_j w_j}\cdot\big(\frac{\theta_i}{1 - \theta_i}\big)^{-\tau_j w_j}\big]^{2}}
    \end{equation}
    
    Simplifying, using the equations (\ref{eq:fo}) and (\ref{eq:r1}), we have:
    
    \begin{equation}
        \label{eq:e_bj}    
        \frac{\partial E}{\partial b_j} = - E^{2}\cdot w_j \cdot \text{ln}[\Phi(\theta_i, \delta_j)] \cdot \big[\Phi(\theta_i, \delta_j)]^{\tau_j w_j} \cdot \frac{\partial \tau_j}{\partial b_j}
    \end{equation}
    
    So, replacing the equation (\ref{eq:e_bj}) in (\ref{eq:ap4}), getting:

    \begin{equation}
            \frac{\partial H}{\partial b_j} = - \sum_{i=1}^{M}\sum_{j=1}^{N} -p_{ij}\cdot E\cdot w_j \cdot \text{ln}[\Phi(\theta_i, \delta_j)] \cdot \big[\Phi(\theta_i, \delta_j)]^{\tau_j w_j} \cdot \frac{\partial \tau_j}{\partial b_j}
    \end{equation}
    
    Finally, we can calculate $\frac{\partial \tau_j}{\partial b_j}$, like this:
    
    \begin{equation}
        \frac{\partial \tau_j}{\partial b_j} = \frac{\partial (\text{tanh}(b_j))}{\partial b_j} = \frac{(e^{b_j} + e^{-b_j})^{2} - (e^{b_j} - e^{-b_j})^{2}}{(e^{b_j} + e^{-b_j})^{2}} =
    \end{equation}
    \begin{equation}
        = 1 - \bigg(\frac{e^{b_j} - e^{-b_j}}{e^{b_j} + e^{-b_j}}\bigg)^{2} = 1 - 
        [\text{tanh}(b_j)]^{2} = 1 - \tau_j^{2}
    \end{equation}
\end{proof}

\end{appendix}

\end{document}